\pdfoutput=1
\documentclass[11pt,a4paper]{article}

\usepackage[T1]{fontenc}
\usepackage[a4paper,margin=2.2cm]{geometry}
\usepackage{graphicx}
\usepackage{xcolor}
\usepackage{amsmath,amssymb,amsfonts,amsthm}
\usepackage{booktabs}
\usepackage{mathrsfs}
\usepackage{multirow}
\usepackage[labelsep=period]{caption}
\usepackage{subcaption}
\usepackage{authblk}
\usepackage{url}
\usepackage[colorlinks,linkcolor=blue,citecolor=blue,urlcolor=blue,breaklinks]{hyperref}

\begin{document}

\title{Continued Pretraining of FinBERT on Finnish Histopathological Reports: Train-Time Signals and Proxy Downstream Correlations}

\author[1,2,3]{Rami Luisto\thanks{Correspondence: Rami Luisto, Email: \href{mailto:rami.m.luisto@jyu.fi}{rami.m.luisto@jyu.fi}.}}
\author[1]{Liisa Pet\"ainen}
\author[1]{Tommi Gr\"onholm}
\author[4]{Jan B\"ohm}
\author[4]{Maarit Ahtiainen}
\author[4]{Tomi Lilja}
\author[1]{Ilkka P\"ol\"onen}
\author[1,5]{Sami \"Ayr\"am\"o}

\affil[1]{Faculty of Information Technology, University of Jyv\"askyl\"a, Jyv\"askyl\"a, Finland}
\affil[2]{Heart and Lung Center, Helsinki University Hospital, Helsinki, Finland}
\affil[3]{Qutwo, Helsinki, Finland}
\affil[4]{Central Finland Biobank, Jyv\"askyl\"a, Finland}
\affil[5]{Wellbeing Services County of Central Finland, Jyv\"askyl\"a, Finland}

\date{}

\maketitle

\begin{abstract}
    In Natural Language Processing (NLP) classification tasks where a lack of labeled data is an issue, continued pretraining (CPT) of transformer models on unlabeled data is an established approach. In this paper, we have two aims. 
    (1) We describe our observations from continued pretraining of the Finnish BERT transformer model (FinBERT) on a Finnish histopathological dataset (below, \emph{the Histopathology data}). 
    (2) Since the Histopathology data has no classification labels, we gather public Finnish datasets as proxy data to analyze whether the signals observed in (1) are associated with downstream classification gains.
    We observe that CPT train-time loss curves differ strongly by domain, and that, in an exploratory analysis, certain CPT-derived features correlate with proxy classification improvement. In particular, this report contributes to the limited literature on NLP for Finnish healthcare data.
\end{abstract}

\noindent\textbf{Keywords:} continued pretraining; FinBERT; Finnish NLP; medical NLP; masked language modeling; representation similarity

\section{Introduction}
\label{sec:introduction}

Our driving motivation here is Natural Language Processing (NLP) work in the healthcare sector, in particular with regard to Finnish medical text data. Finnish is a low-resource language with complex grammatical structures, and is therefore at a disadvantage in the NLP field when compared to larger languages. In particular, off-the-shelf multilingual models or translation-based approaches often fall short \cite{myllyla2025extracting}. Furthermore, a specific feature of any AI work in the healthcare sector is that datasets are delicate and heavily regulated \cite{bani2020privacy, keshta2021security}. It is not uncommon in a healthcare project to end up with only a partial dataset, e.g.\ only text without labels, either because of a mistaken data request or because labeling requires expert medical knowledge. The process of acquiring these labels can then take months of calendar time. In this project, we were interested in seeing whether we could make good use of such waiting time.

\begin{figure}[htbp]
    \centering
    \includegraphics[width=0.9\textwidth]{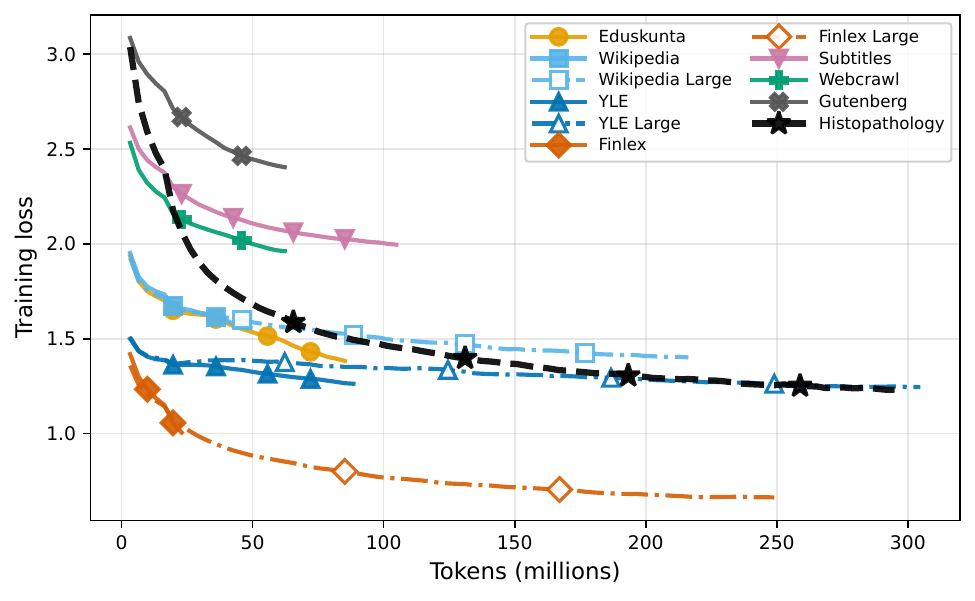}
    \caption{Training loss during continued pretraining of FinBERT on various datasets it is more or less ``familiar'' with. Markers represent epoch boundaries.}
    \label{fig:mlm_loss}
\end{figure}

More specifically, our aim was to do a continued pretraining run on a histopathological dataset, obtained from the Central Finland Biobank, for which we lacked the means to obtain class labels. In this paper, we present our results on such continued pretraining and its effects. For contextualization and hypothesis generation, we ran a similar training with public datasets as well. We discuss in Section \ref{sec:results} what we were able to observe of the models during and after the training; the central observation is in Figure \ref{fig:mlm_loss}. 

Besides continued pretraining, we ran classification tasks on the datasets where labels were available. Our aim was to generate hypotheses on possible correlations between train-time signals and classification performance. We found a few tentative associations, though a replication study is needed.

Modern NLP tends to separate the training of a language model into stages of increasing specialization. Typically, training starts with self-supervised pretraining on a large text corpus followed by possible domain fine-tuning and then task-specific training. Though this paradigm has no singular origin, it was popularized by ULMFiT \cite{howard2018universal}, GPT \cite{radford2018improving}, and BERT \cite{devlin2019bert}. Such ideas were also demonstrated already by Dai and Le \cite{dai2015semisupervised} 
and Peters et al.\ \cite{peters2018elmo}. Regardless, this staged approach is particularly suitable for models based on the transformer architecture, see e.g.\ \cite{vaswani2017attention, devlin2019bert, liu2019roberta, rogers2021primer}. Recent work
has continued to study the benefits of the intermediate adaptation stage: Gururangan et al.\ \cite{gururangan2020don} distinguish
\emph{domain-adaptive} from \emph{task-adaptive} pretraining and show that both approaches can improve downstream performance.
In this paper, we do not make a differentiation at this resolution and just talk of \emph{continued pretraining} (CPT).

While various types of pretraining have been extensively studied for English medical texts (e.g., BioBERT \cite{lee2020biobert}, ClinicalBERT \cite{alsentzer2019publicly}), applying these techniques to minority or low-resource languages can present special challenges. The relative scarcity of healthcare data limits the available tools, and historically multilingual models are not always competitive with language-specific models, see \cite{virtanen2019multilingual}. Recent literature also highlights the necessity of localized, in-domain data for effective adaptation. For instance, Myllyl\"a et al.\ \cite{myllyla2025extracting} demonstrated that when extracting information from unstructured Finnish radiology reports, translating the reports to English to leverage existing models yielded worse results than using native Finnish texts.

Similar observations have been made across other low-resource languages. In Turkish, T\"urkmen et al.\ \cite{turkmen2023bioberturk} found that continuing pretraining on even a small, highly specific in-domain corpus was more effective for clinical text classification than utilizing broader, generic biomedical corpora. For European Portuguese, the MediAlbertina model \cite{nunes2024medialbertina} drastically reduced perplexity and improved information extraction simply by undergoing domain adaptation on local electronic medical records. Likewise, Bui et al.\ \cite{bui2025finetuning} showed that fine-tuning on compiled Vietnamese medical textbook and forum data significantly improved health communication performance, bypassing the heavy ``English bias'' inherent in the base models.

Especially in the realm of Large Language Models (LLMs) it can be crucial to predict from smaller exploratory training runs what the optimal configuration of model size, training-data amount, and compute budget would be. The literature records various power laws that can be used to extrapolate model training behavior. Such power laws exist both for comparing models within a data domain (see e.g.\ \cite{kaplan2020scaling,hoffmann2022training} and references therein) and for comparing behavior across domains, see \cite{brandfonbrener2024loss}. Though such work is typically focused on LLMs instead of Small Language Models (SLMs) like BERT\footnote{BERT technically stands for \emph{Bidirectional Encoder Representations from Transformers}, but it has essentially turned into a lexicalized acronym like laser or radar.} \cite{devlin2019bert}, the underlying transformer architecture is the same, and many of the effects transfer. We will not be fitting our train-time observations to such power laws, but they will provide us with a lens for qualitative analysis of the results.

In this work, our goal is twofold. 
First, we study the effect of CPT on a dataset without training labels and report on how various tools fare in observing such effects. Second, we replicate classical studies on how CPT can affect classification task performance in the setting of the Finnish BERT model. We emphasize that our aim is to observe these phenomena and make qualitative assessments. In particular, we do not aim for optimal classification algorithms as we wish to observe performance differentials rather than absolute levels. Finally, a considerable part of the driving motivation of this work has been the scarcity of reports and research on NLP in the domain of Finnish-language healthcare data. With access to such data being highly regulated, these kinds of reports can prove to be valuable in future work.

\section{Data collection}
\label{sec:data}

Our target model here is the Finnish-language BERT variant introduced by Virtanen et al.\ \cite{virtanen2019multilingual}, \texttt{TurkuNLP/bert-base-finnish-cased-v1}, which we refer to here as \emph{FinBERT}. To discuss CPT, we first need to describe the data used for the original pretraining of FinBERT, as CPT behaves differently depending on whether the model has already seen the continued-pretraining data. (See e.g.\ \cite{prince2023understanding}.)

\subsection{The FinBERT pretraining data}
\label{subsubsec:data-finbert}

The FinBERT model was trained on a corpus of data combining three main sources: news, discussions, and webcrawl. The news data was gathered from the YLE corpus\footnote{\url{http://urn.fi/urn:nbn:fi:lb-2017070501}}
and the STT corpus\footnote{\url{http://urn.fi/urn:nbn:fi:lb-2019041501}}, the discussions were from the Suomi24 corpus\footnote{\url{http://urn.fi/urn:nbn:fi:lb-2019010801}} and the webcrawl data was a combination of \cite{luotolahti2015towards}, a (Finnish) subset of the Common Crawl project\footnote{\url{https://commoncrawl.org}}, and a supplement from Finnish Wikipedia. We refer to \cite{virtanen2019multilingual} for the details of the data gathering and processing, and just replicate their data distribution in Table \ref{table:finBERT-pretrainingdata}.

\begin{table}[t!]
    \centering
    \small
    \begin{tabular}{lrrrr}
    \toprule
               & Docs & Sents & Tokens & Chars \\ \midrule
    News       &   4M &   68M &  0.9B  & 6B  \\ 
    Discussion &  83M &  351M &  4.5B  &  28B  \\ 
    Crawl      &  11M &  591M &  8.1B  & 55B\\ \midrule 
    Total      &  98M & 1\,010M & 13.5B & 89B \\ \bottomrule 
    \end{tabular}
    \caption{FinBERT pretraining text source statistics as reported in \cite{virtanen2019multilingual}.}
    \label{table:finBERT-pretrainingdata}
\end{table}

In particular, we emphasize that the model has not been trained specifically on medical data. We presume the datasets contain some medical text, especially through Wikipedia and webcrawls, but medical text was not a specific focus.

\subsection{Data used in this project}
\label{subsubsec:data-thisproject}

In this project, we used both a private medical dataset of histopathological reports, provided by the Central Finland Biobank and hosted in the Acamedic\footnote{We note that ``Acamedic'' is the name of a secure cloud environment built by the HUS Helsinki University Hospital rather than a misspelling of ``academic''.} secure cloud environment, and a collection of publicly available datasets. The Histopathology dataset contained about 43M tokens' worth of Finnish medical patient records describing both patient visits and clinical sample analysis statements.

As for the publicly available data, the amount of freely available Finnish text data is much smaller than for, e.g., English. In particular, it is not easy to find other large sources not already included in the FinBERT pretraining dataset. We did, however, want to gather a collection of publicly available datasets containing both data that resembles the original pretraining data and data from sources that were unrepresented, or at least underrepresented, in the pretraining dataset. Our collection of ten dataset variants from seven domains plus our Histopathology data is listed in Table \ref{table:dataset_stats} and described in more detail in Appendix \ref{sec:appendix_data}. In the analyses, small and large versions of the same source corpus are treated as separate dataset variants. Unless otherwise noted, each text corpus was split into sets A/B/C in a 70/15/15 distribution. In such a split, set~A was used for CPT training, set~B for CPT evaluation, then set~B again for classification training and set~C for classification testing.

\begin{table}[ht!]
    \centering
\small
\begin{tabular}{lrrrrrr}
\toprule
\textbf{Dataset} & \multicolumn{1}{c}{\textbf{Training}} & \multicolumn{1}{c}{\textbf{Tokens}} & \multicolumn{1}{c}{\textbf{Median}} & \multicolumn{1}{c}{\textbf{Mean}} & \multicolumn{1}{c}{\textbf{$<$ 256}} & \multicolumn{1}{c}{\textbf{Labels}} \\
& \multicolumn{1}{c}{\textbf{Samples}} & \multicolumn{1}{c}{\textbf{(Millions)}} & \multicolumn{1}{c}{\textbf{Tokens}} & \multicolumn{1}{c}{\textbf{Tokens}} & \multicolumn{1}{c}{\textbf{(\%)}} & \multicolumn{1}{c}{\textbf{(classes)}} \\
\midrule
    \textbf{Histopathology} & $\sim$125\,000 & $\sim$30 & n/a & $\sim$243 & n/a & No labels \\
    Eduskunta & 34\,374 & 1 & 24 & 26.6 & 100.0 & 15 \\
    Finlex & 16\,912 & 8 & 1\,213 & 6\,397.5 & 28.8 & 2 \\
    Finlex Large & 162\,627 & 82 & 514 & 510.1 & 0.0 & 2 \\
    Gutenberg & 42\,000 & 21 & 490 & 501.9 & 0.2 & 7 \\
    Subtitles & 41\,262 & 20 & 495 & 488.9 & 2.1 & No labels \\
    Webcrawl & 42\,000 & 12 & 284 & 289.7 & 41.7 & 8 \\
    Wikipedia & 33\,985 & 9 & 241 & 263.0 & 53.5 & 8 \\
    Wikipedia Large & 85\,234 & 22 & 232 & 254.5 & 56.7 & 8 \\
    YLE & 34\,997 & 9 & 237 & 259.0 & 55.7 & 8 \\
    YLE Large & 120\,214 & 31 & 241 & 260.5 & 54.7 & 8 \\
\bottomrule
\end{tabular}
\caption{Training split statistics and token length distribution for each dataset. The $<$ 256 column lists what percentage of samples had fewer than 256 tokens, and the Labels column gives the number of classification classes (``No labels'' where none were available). Histopathology values are approximate as statistics were not recorded at a detailed level during training.}
\label{table:dataset_stats}
\end{table}

Of these datasets, the YLE variants are the only ones whose exact texts we know were included in the FinBERT pretraining. We sampled the Wikipedia data from the November 2023 dump (extraction run in 2025), so it should be at least somewhat different from what FinBERT was shown during its training in 2019. The other datasets, except for the Histopathology data, are ones the FinBERT model \emph{might} have seen at least partially during pretraining through the webcrawled data. Note that the Eduskunta dataset is a strong outlier in that it has very short texts.

\section{Model and methods} 
\label{sec:methods}

Our work in this paper revolves around the FinBERT model. In many cases the more modern SLM variants like RoBERTa \cite{liu2019roberta} or especially ModernBERT \cite{warner2025smarter} would be more suitable for this kind of testing, but for the low-resource language of Finnish, language-specific models have historically been important. Indeed, Virtanen et al.\ \cite{virtanen2019multilingual} showed that a Finnish BERT clearly outperformed the multilingual BERT of its time.
More recent models have narrowed this gap, and the contemporary balance depends on the task, domain, and scale. Even though FinBERT might not be the best-performing model for this task, it provides a natural choice for a base model for observing the effects of CPT. For more details of the BERT architecture we refer the reader to \cite{virtanen2019multilingual, devlin2019bert}.

For the purposes of this paper, we note that FinBERT ingests as input a text string, splits it into a sequence of substrings called \emph{tokens}\footnote{These are typically short common words or syllables.} and processes this sequence into a sequence of so-called \emph{embedding vectors}; for FinBERT these vectors are 768-dimensional. This sequence of embedding vectors then goes through 12 \emph{transformer blocks} to produce a final output sequence of embedding vectors. Crucially, the FinBERT model always prepends a special \texttt{[CLS]} token to any token sequence. The embedding vector of this token aims to capture the semantic content of the whole input string, and it is this embedding vector we mostly work with in this paper. Thus any set of input texts is converted to a set of 768-dimensional \texttt{[CLS]} embeddings, which we call the (embedding) point cloud. These points have natural distance and similarity measures like the Euclidean metric or the cosine similarity of vectors. We also note that the FinBERT model has a maximum ``token window size'' of 512 tokens, meaning that any text longer than that needs to be cut down to 512 tokens.

From the FinBERT core model we train CPT-adapted versions (which we call the \emph{CPT models}) on various datasets, as described in Section \ref{subsec:experiments-mlm}. These CPT models are compared to the base FinBERT model both through unlabeled embedding-vector comparisons and through their performance in a labeled classification task. For the classification task we ran several variants, but for the sake of brevity we limit ourselves to discussing approaches where we trained a simple logistic regression or a $k$-nearest neighbors ($k$-NN) classifier on small subsets of embedding vectors generated by the model; see Section \ref{subsec:results-frozenclass} for details. 

Our training runs were executed on one of two machines: a MacBook Pro (M4 Max, 36 GB RAM) or a Tesla V100-equipped virtual machine running Linux. The various CPT runs usually took several hours each and our classification runs at most a few minutes.

\subsection{Continued pretraining}
\label{subsec:experiments-mlm}

For CPT we use the Masked Language Modeling (MLM) task\footnote{Also known as the \emph{Cloze task} in the more classical literature.} in which the model predicts randomly masked tokens with a cross-entropy loss, see \cite{devlin2019bert}. For each dataset variant (described in Section \ref{sec:data}) we continued training the FinBERT model on the MLM objective with the standard masking probability of 15\% (of the masked tokens, 80\% are replaced by \texttt{[MASK]}, 10\% by a random token, and 10\% are left unchanged). All runs shared the following configuration: AdamW optimizer ($\beta_1 = 0.9$, $\beta_2 = 0.999$, $\varepsilon = 10^{-8}$) with a learning rate of $5 \times 10^{-5}$, linear warmup over the first 500 optimizer steps followed by linear decay, weight decay of $0.01$, gradient-norm clipping at $1.0$, an effective batch size of 64 sequences, a maximum sequence length of 512 tokens (longer inputs truncated, shorter inputs padded to full length), and random seed 42. We evaluated on a validation set every 500 steps and kept the checkpoint with the lowest evaluation loss; no early stopping was used. The number of epochs varied between three and five across datasets; Table~\ref{table:cpt_runs} lists the realized epochs, optimizer steps, and best evaluation losses per dataset. This configuration is similar to what is used in the original FinBERT pretraining \cite{virtanen2019multilingual}, though the original training also used the extra task of Next Sentence Prediction (NSP), which we decided to omit as it seems to provide little extra benefit according to the literature, see \cite{liu2019roberta}.

Most runs used BF16 mixed precision\footnote{In \emph{mixed-precision training} part of the
computation is carried out in 16-bit floating-point numbers instead of the usual 32-bit
ones, which reduces memory use and speeds up training. BF16 and FP16 are two such 16-bit
formats: BF16 spans the same range of magnitudes as a 32-bit float but with fewer
significant digits, while FP16 is more precise over a narrower range. Which of the two is
available depends on the hardware.} on the M4 Max machine with a per-device batch size of 16 and gradient accumulation over 4 batches. The Finlex, Finlex Large, and Gutenberg runs were instead executed on the V100 machine, which lacks hardware BF16 support, using FP16 mixed precision, gradient checkpointing, and smaller per-device batches with correspondingly larger gradient accumulation. The effective batch size of 64 was identical across all runs. The Histopathology run, executed inside the Acamedic environment, used the same hyperparameters as the M4 Max runs. 

\begin{table}[ht!]
    \centering
    \small
    \begin{tabular}{lrrrr}
    \toprule
    \textbf{Dataset} & \textbf{Train} & \textbf{Epochs} & \textbf{Optimizer} & \textbf{Best} \\
     & \textbf{samples} & & \textbf{steps} & \textbf{eval loss} \\
    \midrule
    Histopathology  & $\sim$125\,000 & 5 & $\sim$9\,000 & 1.168 \\
    Eduskunta       & 34\,374  & 5 & 2\,690 & 1.403 \\
    Finlex          & 16\,912        & 3 & 795    & 0.884 \\
    Finlex Large    & 162\,627       & 3 & 7\,626 & 0.595 \\
    Gutenberg       & 42\,000        & 3 & 1\,971 & 2.281 \\
    Subtitles       & 41\,262        & 5 & 3\,225 & 1.902 \\
    Webcrawl        & 42\,000        & 3 & 1\,971 & 1.858 \\
    Wikipedia       & 33\,985        & 3 & 1\,596 & 1.453 \\
    Wikipedia Large & 85\,234        & 5 & 6\,660 & 1.335 \\
    YLE             & 34\,997        & 5 & 2\,735 & 1.208 \\
    YLE Large       & 120\,214       & 5 & 9\,395 & 1.176 \\
    \bottomrule
    \end{tabular}
    \caption{Realized CPT training runs per dataset variant. Histopathology values are approximate as exact numbers were not recorded during training.}
    \label{table:cpt_runs}
\end{table}

\subsection{Measuring embedding changes}
\label{subsec:MeasuringChanges}

Our core idea was to quantify how much CPT affected a model by measuring changes in text embedding vectors. In particular, after running CPT with our training data (set~A), we extracted the \texttt{[CLS]} embedding vectors of the validation dataset (set~B) from each of FinBERT's 1+12 transformer layers\footnote{The BERT architecture has 12 \emph{transformer blocks} or \emph{layers}. We denote the collection of context-free embeddings for the input tokens as layer 0 embeddings. See again \cite{virtanen2019multilingual} or \cite{devlin2019bert} for details.}. Thus from the validation dataset we created 13 embedding point clouds from both the base and CPT models, which we denote $E^{\text{base}}_{0},\ldots, E^{\text{base}}_{12}$ and $E^{\text{CPT}}_{0},\ldots, E^{\text{CPT}}_{12}$. We then observed how much the point clouds $E^{\text{base}}_j$ and $E^{\text{CPT}}_j$ resembled each other for each $j = 0, \ldots, 12$.

The following techniques are common tools for measuring the similarity of two point clouds. Instead of providing detailed definitions, we give brief descriptions and refer the reader to standard sources. See \cite{ding2021grounding, wu2025measuring, bo2024evaluating} for further comparative studies on how they can be used in analyzing neural networks in general, and how they (and other approaches) differ. See also \cite{williams2021generalized} for a more foundational approach to measuring ``neural representations'' in general.

\begin{itemize}
    \item \textbf{Centered Kernel Alignment (CKA)}: A rotation-invariant similarity metric that compares the geometric structure of two sets of vectors based on the distribution of cosine similarities within each set. \cite{kornblith2019similarity, maheswaranathan2019universality, phang2021fine}

    \item \textbf{Procrustes Analysis}: Measures the optimal alignment between two shapes. We compute the Procrustes distance (or similarity) after finding the best orthogonal rotation to map the CPT embeddings to the base embeddings. \cite{smith2017offline, haxby2011common}

    \item \textbf{Representational Similarity Analysis (RSA)}: Compares the correlation structures of the two spaces. We calculate the distance matrices of both spaces and then compute the correlation between these matrices. \cite{dwivedi2019representation, kriegeskorte2008representational}
\end{itemize}

We also measured the internal geometry of the point clouds $E^{\text{base}}_{12}$ and $E^{\text{CPT}}_{12}$. The following approaches measure each point cloud separately and then compare the results.
\begin{itemize}
    \item \textbf{Isotropy}: We use the \emph{Partition Function $Z(\tau)$} and \emph{Effective Rank} as metrics to measure how uniformly the embeddings occupy the latent space. \cite{mu2017all,roy2007effective, rudman2022isoscore}
    
    \item \textbf{Clustering}: We applied $k$-means clustering, choosing $k$ by the Silhouette score, and, where class labels were available, quantified how well the clusters align with the labels using the Adjusted Rand Index (ARI) and Normalized Mutual Information (NMI). \cite{strehl2002cluster, rousseeuw1987silhouettes, hubert1985comparing}
\end{itemize}

\subsection{Classification tasks}
\label{sec:classification}

The typical approach when using BERT-based models for classification, as already advocated in \cite{devlin2019bert}, is to fine-tune the whole model on the classification task. Indeed, the ``raw'' \texttt{[CLS]} embeddings are known to capture semantic content poorly, see e.g.\ \cite{li2020sentence} and references therein. Even so, we deliberately depart from this approach, opting instead to train simple classifiers on the frozen \texttt{[CLS]} embeddings. Our core rationale is that our aim is not to evaluate the \texttt{[CLS]} embedding quality, nor to optimize the classification results, but to observe the effects of CPT. In an exploratory stage of the experiments we did indeed run a few ``full'' classification training runs, but once the whole model is trained on the task, the base and CPT models converge to similar performance and the CPT effect is largely washed out.

Furthermore, both to simulate limited-data situations and to enhance the signal, we limited our classification training-set size. In particular, for our classification task we sampled a limited subset (e.g.\ 500, 1\,000, or 2\,000 samples) of set~B. For classifiers we used either a simple logistic regression or a $k$-NN classifier. 

We trained the classifiers on the raw 768-dimensional final-layer
\texttt{[CLS]} embeddings without any normalization, standardization, or
dimensionality reduction. For logistic regression we used L2
regularization, the LBFGS solver, a maximum of 2\,000 iterations, and
inverse class-frequency weights to counter class imbalance; the
regularization strength $C$ was selected by 5-fold cross-validated grid
search over 20 logarithmically spaced values in $[10^{-4}, 10^{4}]$,
optimizing macro-F1. For the $k$-NN classifier we likewise selected the
hyperparameters by stratified 5-fold cross-validated grid search
(optimizing macro-F1) over $k \in \{3, 5, 7, 11, 15, 21\}$, uniform versus
distance weighting, and cosine versus Euclidean distance, independently for
every combination of dataset, training-subset size, random seed, and model
(base or CPT). Each experiment drew its labeled subset of the training data with
class-proportional allocation, trained on it, and evaluated on a held-out dataset; we repeated every experiment with five random seeds (42,
123, 456, 789, 1024) and report means over the repetitions. The same
sampled subsets were used for the base and the CPT model, so the comparison
between the two is paired.

Our datasets had widely differing label counts, label balances, and general data quality. Simply comparing e.g.\ the increase in accuracy from CPT for a given classification task would bias the comparison toward domains with lower baseline accuracy. This is because increasing a classifier's accuracy from 90\% to 95\% is much harder than from 40\% to 45\% -- not to mention the fact that from the starting point of 50\% one can improve by 15 percentage points, but starting from 92\% one cannot. Thus we had a few different measures that we used for both accuracy and F1 scores. Here $s_{\mathrm{CPT}}$ stands for the accuracy or F1 score of the CPT model, while $s_{\mathrm{BL}}$ stands for the baseline score of the base model.
\begin{enumerate}
    \item Error Reduction Rate (ERR): what fraction of remaining error was removed
    $(s_{\mathrm{CPT}} - s_{\mathrm{BL}})/(1-s_{\mathrm{BL}})$.

    \item Logit delta: $\operatorname{logit}(s_{\mathrm{CPT}}) - \operatorname{logit}(s_{\mathrm{BL}})$, where $\operatorname{logit}(p) = \log(p/(1-p))$.
\end{enumerate}

\section{Results}
\label{sec:results}

We discuss our results across dataset variants in parts: CPT training, the classification task, and then the observed correlations between CPT and classification performance. We note that the Eduskunta dataset was an outlier in many measurements, likely because its texts are very short; see again Table \ref{table:dataset_stats} for the distribution. Note that text length can have strong effects on NLP learning \cite{nagatsuka2023length}.

\begin{figure}[htbp]
    \centering
    \includegraphics[width=0.9\textwidth]{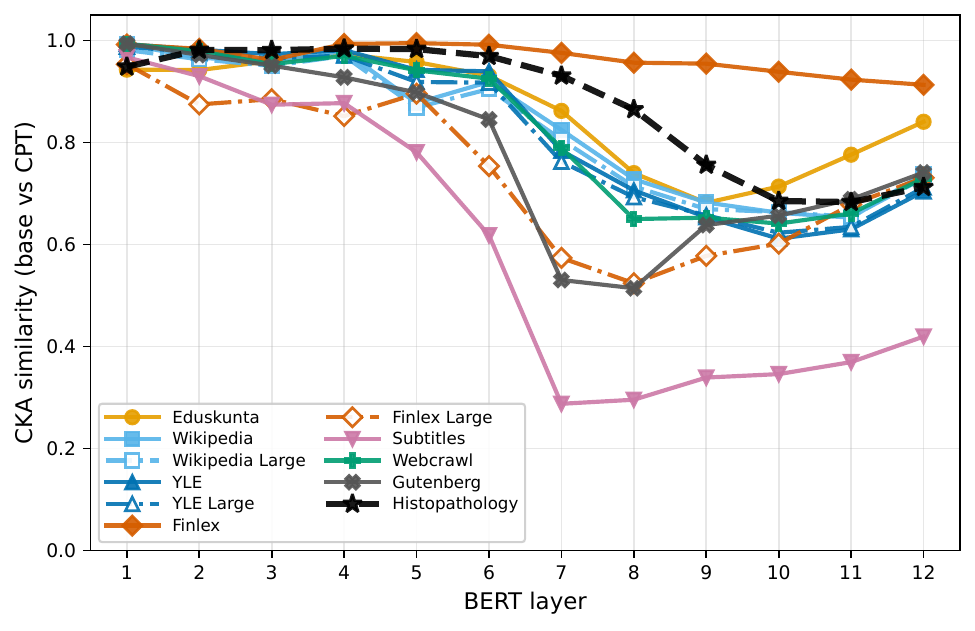}
    \caption{CKA similarity between the base and CPT models across FinBERT layers, computed on the \texttt{[CLS]} embeddings of set~B; higher values indicate greater similarity. Layer 0 (the context-free token embeddings) is omitted as the CKA scores were degenerate.}
    \label{fig:results-CKA}
\end{figure}

\subsection{Continued pretraining}
\label{subsec:results-mlm}

In Figure \ref{fig:mlm_loss} we show the train-time loss curves of the CPT. What we find most striking is that the two YLE datasets show very little change in the loss, while the Finlex and Histopathology datasets show a drastic change. We interpret this as evidence that the YLE data was already ``familiar'' to the FinBERT model and produced very little new adaptation need, while the datasets that were more novel to the model, like Histopathology or Finlex, caused drastic changes.

This type of interpretation can be studied through the lens of the scaling laws used to predict LLM train- and test-time behavior. The various classical sources, e.g.\ \cite{kaplan2020scaling,hoffmann2022training, brandfonbrener2024loss},
advocate for slightly different variations of power laws. For the sake of simplicity, we view our findings through the one used by Hoffmann et al.\ in \cite{hoffmann2022training}. It describes the power law\footnote{We emphasize here that we are not aiming to fit a power law to our data, but rather use this as a tool for a qualitative analysis.} for the pretraining loss $L$ as
\begin{align}
    L(N, D) = E + \frac{A}{N^\alpha} + \frac{B}{D^\beta}
\end{align}
where $E$ is the underlying entropy of the text domain from which the training data is sampled (see e.g.\ \cite{mackay2003information}), $N$ is the number of model parameters, $D$ is the number of training tokens, and the rest are constants that depend on the model and data context. The term $AN^{-\alpha}$ describes the \emph{approximation error} we get from fitting a model with limited parameters and capability to our data, whereas the term $BD^{-\beta}$ describes the \emph{estimation error} we get from having only a small sampling of all possible data.

Since we are working with a fixed model, we do not vary $N$, and so for us the power law takes the form $L(D) = C + B D^{-\beta}$. With this as our theoretical framework we can suggest an explanation for the shape of the loss curves in Figure \ref{fig:mlm_loss}. The horizontal asymptotes of the train loss curves represent a combination of the underlying entropy of the particular dataset and the approximation error we get from using FinBERT rather than a perfect oracle model. 

To analyze the magnitude of the loss-curve \emph{drop}, it is important to note that the original power laws were built for the \emph{pretraining} stage and not for CPT. In particular, the almost flat shapes of the training loss curves of the YLE datasets probably reflect the fact that the FinBERT model has already seen the data. We are essentially seeing here the flatter tail end of a larger loss curve that ``started'' during the FinBERT pretraining. For the other domains, especially Finlex and Histopathology, the very strong drop indicates that such data was most likely not present in the pretraining data, and the magnitude of the drop is then a sign of how much estimation error we were able to reduce.

CKA, RSA, and Procrustes measured how the \texttt{[CLS]} embedding point clouds of the set~B texts differed between the CPT and base models; the results were very similar across the three methods. Figure \ref{fig:results-CKA} displays the CKA result. This should reflect not only the change in the weights of the model due to CPT, but more crucially how the model's ``understanding'' of this domain's text content changed.

We note that the changes do not seem to directly correlate with our train-time loss dynamics. In Figure~\ref{fig:results-CKA}, the Subtitles dataset is almost an outlier, showing very large changes, while the Finlex dataset shows the least change. There seems to be a non-linear connection between how much the model adapted to the text and how much change in the embedding vectors the learning required. We note, however, that these results are highly dependent on the text content of the particular domain; we are not probing the model layers directly but rather the embeddings of texts in set~B.

\subsection{Classification task}
\label{subsec:results-frozenclass}

\begin{figure}[htbp]
    \centering
    \includegraphics[width=0.9\textwidth]{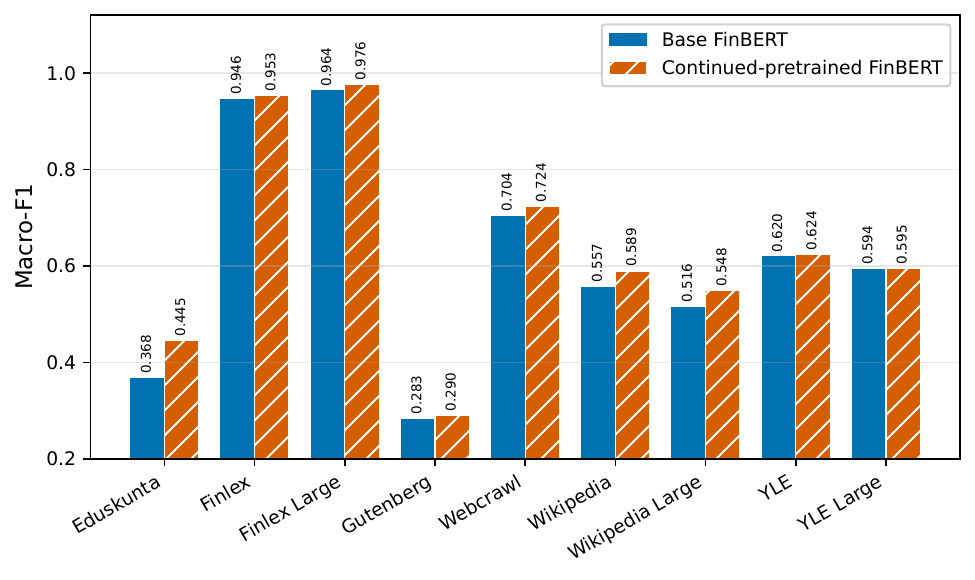}
    \caption{Macro-F1 of the base and CPT models for the logistic regression classifier across the labeled dataset variants, trained on 500 labeled samples.}
    \label{fig:results-scarce-comparison}
\end{figure}

Figure \ref{fig:results-scarce-comparison} shows one of our classification macro-F1 results. We note that the classification macro-F1 scores vary considerably across dataset variants. To demonstrate how raw accuracy- and macro-F1-score deltas differ from the relative versions discussed in Section \ref{sec:classification}, we have listed the raw deltas next to logit deltas and ERR scores in Table \ref{table:classification_deltas}.

The results are generally as expected. The CPT training tended to improve the classification results, though with the YLE datasets the effects were close to zero. Once again, the YLE datasets were part of the original FinBERT training data. We note that the different datasets are quite heterogeneous with respect to their amenability to classification.

\begin{table}[ht!]
    \centering
    \footnotesize
    \setlength{\tabcolsep}{4pt}
    \begin{tabular}{lrrrrrrrrr}
    \toprule
     & Eduskunta & Finlex & Finlex & Gutenberg & Webcrawl & Wikipedia & Wikipedia & YLE & YLE \\
     &  &  & Large &  &  &  & Large &  & Large \\
    \midrule
    Accuracy delta & \textbf{0.074} & 0.005 & 0.004 & 0.006 & 0.019 & 0.035 & 0.031 & 0.007 & 0.004 \\
    Accuracy logit delta & 0.296 & 0.146 & \textbf{0.410} & 0.031 & 0.091 & 0.145 & 0.125 & 0.029 & 0.016 \\
    Accuracy ERR & 0.143 & 0.132 & \textbf{0.334} & 0.009 & 0.061 & 0.082 & 0.069 & 0.018 & 0.010 \\
    \midrule
    Macro-F1 delta & \textbf{0.077} & 0.007 & 0.012 & 0.007 & 0.019 & 0.032 & 0.033 & 0.005 & 0.001 \\
    Macro-F1 logit delta & 0.321 & 0.149 & \textbf{0.430} & 0.032 & 0.094 & 0.131 & 0.131 & 0.019 & 0.004 \\
    Macro-F1 ERR & 0.122 & 0.132 & \textbf{0.341} & 0.009 & 0.065 & 0.073 & 0.067 & 0.012 & 0.003 \\
    \bottomrule
    \end{tabular}
    \caption{Accuracy- and macro-F1-based improvement measures for the
    logistic-regression classifier trained on 500 labeled samples: the delta,
    the logit delta, and the error reduction rate (ERR) between the CPT and
    base models, computed from the per-domain mean scores over the five random
    seeds. The largest value in each row is shown in bold.}
    \label{table:classification_deltas}
\end{table}

\subsection{MLM--classification correlations}
\label{subsec:results-mlmvsclass}

Here we discuss correlations between signals observed during CPT and performance changes in the classification tasks between models. We emphasize that the analysis is exploratory and post hoc. In particular, for the sake of transparency we note that \emph{we did not record beforehand which features and targets to use in the comparison}. Indeed, we found some features that seem to correlate with better classification performance, but these were selected from a set of 91 features and 129 targets\footnote{The full heatmap of correlations between all of the features and targets we had is available online at \url{https://github.com/ramiluisto/CPT_analysis_companion/blob/main/figures/heatmap.pdf}.}. We did not apply a correction for multiple comparisons and therefore treat the observed correlations as hypothesis-generating patterns rather than confirmatory evidence. 

\begin{figure}[htbp]
    \centering
    \includegraphics[width=0.9\textwidth]{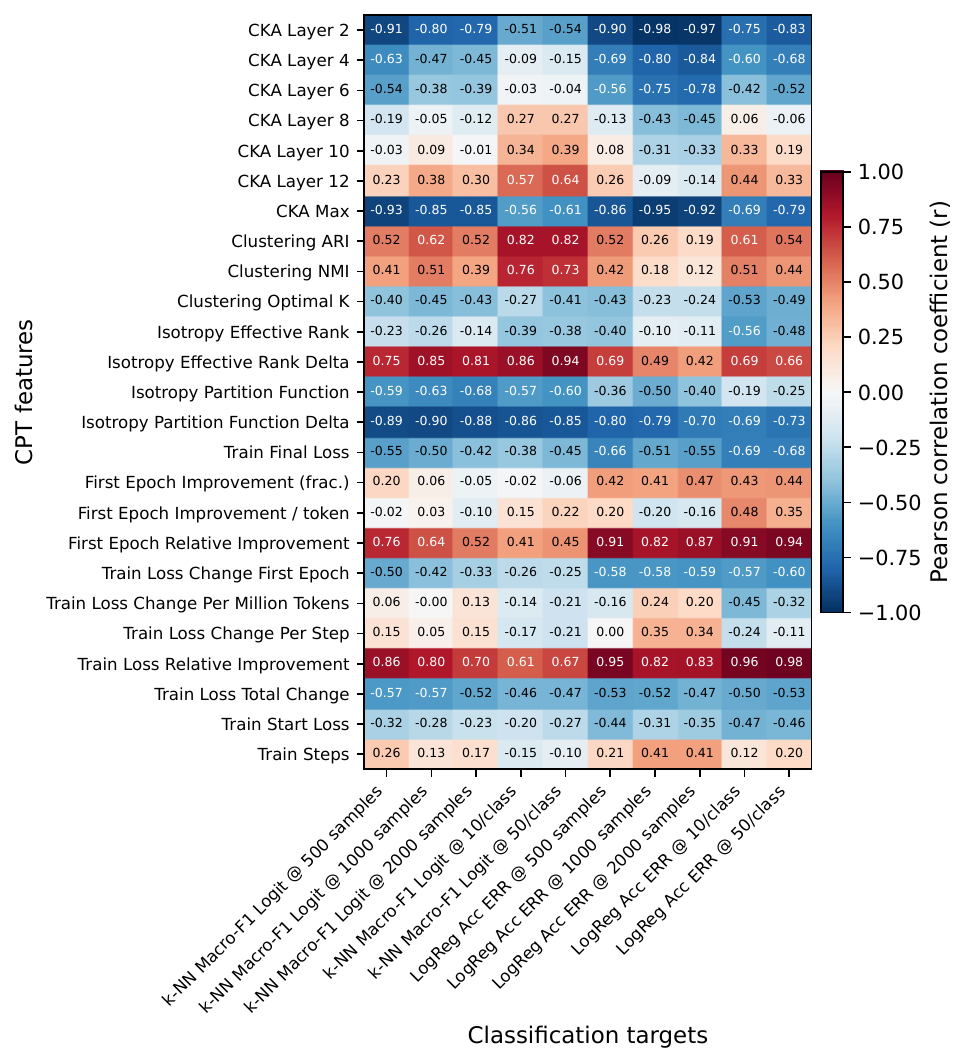}
    \caption{Subset of the correlation heatmap between CPT features and classification task targets across labeled dataset variants.}
    \label{fig:results-heatmap}
\end{figure}

\begin{figure}[htbp]
    \centering
    \includegraphics[width=0.9\textwidth]{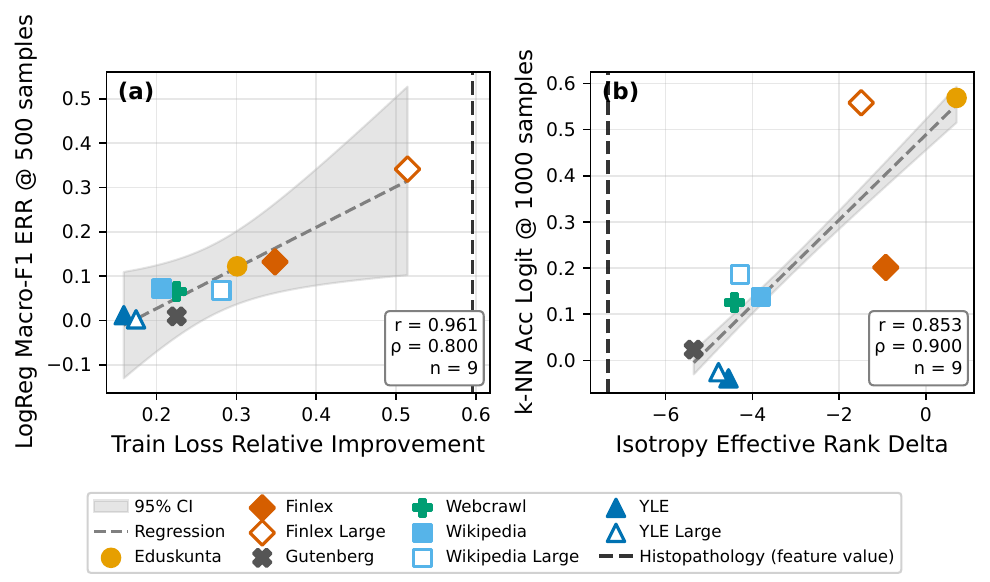}
    \caption{Example correlations between CPT signals and classification improvement. (a) Train-loss relative improvement vs.\ the logistic-regression macro-F1 error-reduction rate (500 training samples). (b) Isotropy effective-rank delta vs.\ the $k$-NN accuracy logit delta (1\,000 training samples). The vertical dashed lines mark the Histopathology dataset's feature values; they are feature-side references only, as no classification labels exist for that dataset.}
    \label{fig:scatter_3x2}
\end{figure}

In Figure \ref{fig:results-heatmap} we show a subset of this heatmap with our most relevant features and targets. Each cell corresponds to a scatter plot of the nine labeled dataset variants -- with the given feature and target as the $x$- and $y$-axes -- and a line fitted to it. (See e.g.\ Figure \ref{fig:scatter_3x2}.) The colors encode Pearson correlation coefficients, with dark red indicating strong positive correlation and dark blue indicating strong negative correlation.

From this heatmap of correlations, we note that there are isolated large-magnitude cells, which we presume to reflect spurious correlations. But there are several strong horizontal strips which seem to persist over different target measures. These are, in an approximate order of prominence:
\begin{enumerate}
    \item \emph{Train loss relative improvement}. This is simply the drop in the CPT train-time loss divided by the loss at the start. (See again Figure \ref{fig:mlm_loss}.)
    \item \emph{Isotropy effective rank delta and Isotropy partition function delta.} In particular, the more isotropic (more evenly distributed) the embeddings became, the better the classifier performed.
    \item \emph{Train first epoch loss relative improvement.} Same as the first point, but restricted to just the first epoch. It seems to be less descriptive than the full relative loss change, though still prominent.
    \item \emph{CKA Max and CKA Layers 1--3.} This also persists for the RSA and Procrustes measures. Perhaps large changes in the earlier layers of the model are an indicator that we had a lot of data for training and thus the changes propagated strongly to the earlier layers as well?
    \item \emph{Clustering ARI.} When clustering improves during CPT, it seems that the clusters tend to be more supportive of our classification tasks, regardless of whether we are using $k$-NN or logistic regression.
\end{enumerate}

Separately from the multiple-comparisons issue, it is not clear whether these different features are correlated or whether they all reflect a common ``considerable deep changes during CPT'' confounder. There could be several distinct effects at play. We also note that the amount of change in the later layers of the model does not seem to strongly correlate with classification performance, which we find curious.

\section{Conclusions, limitations, and future work}
\label{sec:conclusions}

In this paper, we set out to 
(1) measure the effects of CPT on the FinBERT model with histopathological reports and
(2) test, on public proxy datasets, whether any CPT signals correlate with downstream classification performance.
For (1) we observed clearly differing train-time loss-curve behavior between the Histopathology data, which is certainly novel to the model, and the YLE data, which was present in the FinBERT pretraining data.
For (2) our exploratory analysis demonstrated tentative associations between CPT signals and downstream performance. 
In particular, the connection between the different features extracted from the training, especially loss curve geometry, layerwise embedding changes, and embedding isotropy, is suggestive.

None of our tentative findings are statistically established, and they were in line with expectations. We nevertheless consider this work a useful contribution to the sparse literature on Finnish healthcare NLP. At minimum, we provide informed priors for planning CPT when labeled data is lacking.

For future work we see both theoretical possibilities and further practical experiments. On the theoretical side, these types of analyses would also benefit from a mathematical foundation for ``capped power laws for CPT'' that would account for the fact that a model has already seen large amounts of data. Work in this direction already exists, see e.g.\ the transfer scaling laws in \cite{hernandez2021transfer} or the D-CPT law for domain-continual pretraining in \cite{que2024dcpt}. We feel there is room for exploration and advances in this area, especially in the domain of low-resource languages and small models.

On the practical side, another way to obtain CPT signals would be to apply the measures described here to the model weights, neural activations, or attention patterns rather than embedding vectors, see e.g.\ \cite{eilertsen2020classifying, raghu2017svcca, morcos2018insights} for more involved approaches. Replication of our results with other datasets would also be crucial.

Finally, we note that we have omitted from this report some experiments we did not find particularly informative. For the MLM task we also plotted various Principal Component Analysis (PCA), Independent Component Analysis (ICA), t-distributed Stochastic Neighbor Embedding (t-SNE), and Uniform Manifold Approximation and Projection (UMAP) projections of the resulting \texttt{[CLS]} embedding point clouds of the set~B texts. There were visually clear differences in the images between the base and CPT models, but nothing we found worth reporting.

\section*{Acknowledgments}

We gratefully acknowledge the Central Finland Biobank. We thank the Wellbeing Services County of Central Finland for research services. We also thank members of the GenAID research group for medical expert consultation on oncological matters.

This study is one part of the Central Finland AI hub II project that has received funding from the Regional Council of Central Finland (\url{https://www.keskisuomi.fi/}) and the European Regional Development Fund (ERDF) (\url{https://ec.europa.eu/regional_policy/funding/erdf_en}).

Parts of this work were supported by the Strategic Research Council at the Research Council of Finland (grant number 372591) under GAINS
(``Generative AI and Digital Solutions: Enhancing Effectiveness and
Productivity of Healthcare Services'') and by Business Finland (1562/31/2024) through the GenAID research project.

\section*{Use of AI tools declaration}

Throughout this project the first author has used various AI tools, namely Cursor, Claude Code, and ChatGPT, to assist in some of the more mundane parts of programming. Most scripts have been created in collaboration with coding agents, and AI tools have been used for critical feedback on the manuscript. Any work by AI tools has been validated by a human.

The Histopathology dataset resided in the Acamedic environment, and no AI system besides the FinBERT model trained there interacted with the data.

The textual content of this article was written by humans, and the authors take full academic responsibility for its contents.

\section*{Data and code availability}

The public corpora used in this study are available from their original distributors: Finnish Wikipedia (Wikimedia dumps), the YLE Finnish News Archive (Kielipankki, restricted license), Finlex Open Data (Ministry of Justice), the Eduskunta-VKK dataset (GitHub), Project Gutenberg (Gutendex), OpenSubtitles (OPUS/Kielipankki), and \texttt{Finnish-NLP/mc4\_fi\_cleaned} (HuggingFace). Our extraction, preprocessing, and split-construction scripts, together with the configuration files and random seeds needed to reconstruct our datasets and train/validation/test splits from these sources, are published as a public code release accompanying this paper at \url{https://github.com/ramiluisto/CPT_analysis_companion}.
The YLE corpus license does not permit redistribution of derived text data; for that corpus the release provides the scripts and sample identifiers needed to reconstruct our datasets from the original Kielipankki distribution. The Histopathology dataset contains sensitive patient information and cannot be shared; it was accessed and processed exclusively within the Acamedic secure environment.

\section*{Ethics statement}

The Histopathology dataset was obtained from the Central Finland Biobank under a biobank data agreement, in accordance with the Finnish Biobank Act. The data was pseudonymized and was accessed and processed solely within the Acamedic secure computing environment; no patient-level data left the environment, and only aggregate training statistics were exported.

\section*{Conflict of interest}

All authors declare no conflicts of interest in this paper.

\bibliographystyle{AIMS}
\bibliography{bibliography}

\appendix

\section{Details on our datasets}
\label{sec:appendix_data}

All corpora were cleaned by removing samples with missing text, exact
duplicate texts, and texts shorter than 50 characters. Unless stated
otherwise, each corpus was then split into sets A/B/C with ratios 70/15/15
(stratified by class label where labels exist) using a fixed random seed
(42); token counts throughout refer to the FinBERT tokenizer.

\begin{enumerate}
    \setlength{\itemsep}{6pt}

    \item \textbf{Histopathological reports:} This was a dataset provided by the Central Finland Biobank, containing medical descriptions of patients who underwent histopathological examination. Labels were not readily available\footnote{With expert advice it might have been possible to extract some labels from the data, but not necessarily at scale.}, so we used this dataset purely for continued pretraining.
    
    \textbf{Size \& Splits:} Approximately 179\,000 samples of text, with an average length of 243 tokens (80\% of the samples had between 30 and 500 tokens).
    
    \textbf{Labels:} None available.
    
    \textbf{Reference:} Central Finland Biobank (Keski-Suomen Biopankki). Histopathology Reports -- private medical dataset (unpublished, used with permission for this study). Central Finland Biobank, Jyväskylä, Finland. (No public access available.)

    \item \textbf{YLE:} We used the same YLE corpus as in the FinBERT pretraining, and sampled both a small and a large variant of that data. This is data that the FinBERT model has already seen in the pretraining, but a very small part of the full 13.5B tokens used by the FinBERT authors.\footnote{The authors of FinBERT \cite{virtanen2019multilingual} state that the YLE dataset was about 20\% of the news section of their corpus, which would put the full YLE data at about 1.3\% of their pretraining data.} In particular, continuing the pretraining of the FinBERT model on this dataset should ``teach'' the model little that is new, but perhaps narrow its focus.
    
    \textbf{Size \& Splits:} Small variant: $\sim$50\,000 samples (35k train / 7.5k val / 7.5k test). Large variant: 172\,000 samples (120k train / 26k val / 26k test).
    
    \textbf{Labels:} 8 different topical classes (e.g., Politics, Economy, Sports, Culture).
    
    \textbf{Reference:} \cite{yle2017newsarchive}.

    \item \textbf{Wikipedia:} We sampled two datasets from the HuggingFace distribution of the Finnish Wikipedia dump of November 1, 2023. The Wikipedia data is something that the FinBERT pretraining also used, but we note that we drew a fresh sample. In all likelihood the data has at least drifted from the specific text that FinBERT saw during the pretraining.
    
    \textbf{Size \& Splits:} Small variant: 49\,000 samples (34k train). Large variant: 122\,000 samples (85k train).
    
    \textbf{Labels:} 8 topic classes assigned by a keyword-matching heuristic over title and text (7 topics like \emph{Science} or \emph{Politics}, plus one catch-all \emph{Other}). We note that these categories are not Wikipedia's own metadata.

    \textbf{Reference:} \cite{wikimedia2023fiwiki}.

    \item \textbf{Finlex:} We extracted from the Finlex database Finnish legal documents (legislation and international treaties). This is again text that the pretraining might have seen through the webcrawl.
    
    \textbf{Size \& Splits:} Small variant: roughly 24\,000 samples (17k train), with full documents kept intact (average length $\sim$6\,000 tokens). Large variant: 232\,000 samples (162k train) created by sampling disjoint 512-token blocks from the source documents.
    
    \textbf{Labels:} 2 document-type classes: treaty vs.\ statute (ratio $\sim$1:3.4).
    
    \textbf{Reference:} \cite{finlex2025opendata}.

    \item \textbf{Eduskunta:} The Finnish parliament stores records of written questions and answers. We retrieved a dataset of these records for our experiments.
    
    \textbf{Size \& Splits:} 55\,106 sentences in the official splits (49\,106 train / 3\,000 dev / 3\,000 test); the MLM training set is a label-stratified 70\% subset (34\,374 sentences) of the official training split.
    
    \textbf{Labels:} 15 responsible ministries (e.g., Ministry of Finance, Ministry of Education).
    
    \textbf{Reference:} \cite{ajanki2019eduskunta}.
    
    \item \textbf{Subtitles:} The OpenSubtitles project provides a dedicated collection of Finnish subtitles. We sampled a dataset from this project, which was used for CPT experiments only.
    
    \textbf{Size \& Splits:} 59\,000 samples (41k train).
    
    \textbf{Labels:} None (we were not able to find reasonable classes for them).
    
    \textbf{Reference:} \cite{lison2016opensubtitles}. (Finnish subtitles extracted via OPUS.)
    
    \item \textbf{Gutenberg:} We scraped Finnish books from Project Gutenberg.
    
    \textbf{Size \& Splits:} 60\,000 samples (42k train).
    
    \textbf{Labels:} 7 genres: Fiction, Drama, Poetry, Folklore, Humor, Non-fiction, and Children's literature.
    
    \textbf{Reference:} \cite{gutenberg2025finnish}.

    \item \textbf{Webcrawl:} We used an existing Finnish webcrawl dataset from Toivanen et al.
    
    \textbf{Size \& Splits:} 60\,000 samples (42k train), about 12M tokens.
    
    \textbf{Labels:} 8 source-type classes derived from the URL host (News, Government, Educational, Commercial, Forum, Blog, Wikipedia, Other).
    
    \textbf{Reference:} \cite{toivanen2022mc4fi}.

\end{enumerate}

\end{document}